\documentclass[letterpaper]{article} 
\usepackage{aaai2027}  
\usepackage[hyphens]{url}  
\usepackage{graphicx} 
\usepackage{natbib}  
\usepackage{caption} 
\usepackage{subcaption}
\usepackage{algorithm}
\usepackage{algorithmic}
\usepackage{amsmath}
\usepackage{amssymb}

\usepackage{newfloat}
\usepackage{listings}
\DeclareCaptionStyle{ruled}{labelfont=normalfont,labelsep=colon,strut=off} 
\floatstyle{ruled}
\newfloat{listing}{tb}{lst}{}
\floatname{listing}{Listing}
\usepackage{multirow}
\usepackage{booktabs}
\title{SIGMA: Structured Noise-Effect-Aware Grouped Multi-Agent Aggregation}
\author{
    Mingqian Li
}

\affiliations{
    School of Computer Science and technology, Tongji University\\
    Shanghai, China\\
    mingqianli071@gmail.com
}

\begin{document}

\maketitle

\begin{abstract}
Cooperative multi-agent reinforcement learning (MARL) faces significant challenges in maintaining robust coordination under noisy observations. 
Although observation disturbances are often introduced independently across agents, their downstream effects on cooperative decision-making can become structured through underlying cooperation structures. 
We characterize this phenomenon as \emph{structured noise effects}, where noise-induced decision effects exhibit local correlation among agents with stronger task-related dependencies while remaining globally heterogeneous across different agents and local structures. Existing robust MARL methods, however, rarely explicitly characterize or exploit such structure-dependent noise effects. To address this limitation, we propose SIGMA, a hierarchical collaboration framework that exploits cooperation structures to learn robust representations under noisy observations. 
SIGMA first organizes agents into adaptive local structures through density-based grouping and performs intra-group consensus aggregation to preserve shared task-relevant information while smoothing agent-specific representation deviations. 
Inter-group attention then adaptively integrates information across different groups to preserve global coordination while accommodating their heterogeneous contributions. Experiments on noisy-observation tasks in StarCraft II empirically validate the structured noise effects and demonstrate that SIGMA consistently improves robustness under observation noise while maintaining competitive performance in noise-free environments.
\end{abstract}


\section{Introduction}

Cooperative multi-agent reinforcement learning (MARL) has achieved significant progress in solving complex tasks that require coordination among multiple agents, with successful applications in multi-agent games \cite{samvelyan2019starcraft}, multi-robot coordination \citep{sadhu2018efficient,hu2023neuronsmae}, and autonomous driving \cite{li2023conditional}. 
However, most existing MARL methods assume that agents can obtain reliable observations during decision-making \citep{rashid2018qmix,yu2022surprising,wen2022multi}, an assumption that may not hold in real-world deployments where sensing errors, communication limitations, and environmental disturbances are inevitable \cite{muratore2019assessing}. 
Noisy observations can distort agents' local perceptions and subsequently affect their decisions, potentially disrupting coordination and degrading overall team performance \citep{li2022off,liu2022robustness,zhang2020robust,yang2023cooperative}.

Noisy local observations can prevent agents from obtaining reliable information about the environment, thereby hindering cooperative policy learning \cite{muratore2019assessing}. 
Beyond individual perception errors, observation uncertainty can further interfere with cooperative interactions, as unreliable local information may affect how agents coordinate with their teammates \cite{kilinc2018multi}. 
More importantly, the resulting influence is not necessarily confined to the agents whose observations are directly perturbed; perturbing only a single agent can substantially degrade the performance of the entire cooperative team \cite{lin2020robustness}. 
Such team-level effects are also non-uniform across agents, as comparable observation impairments can lead to different cooperative outcomes depending on which agents are affected \cite{barta2025measuring}.

We characterize these non-independent impacts as \emph{structured noise effects}, which exhibit two complementary properties: local correlation and global heterogeneity. 
At the local level, agents with stronger task-related dependencies tend to exhibit more correlated noise-induced decision impacts, reflecting the coupling of disturbance effects through local cooperative interactions. 
At the global level, the magnitude and pattern of these impacts can vary substantially across agents and local cooperation structures, resulting in heterogeneous disturbance responses across the team. 
Together, these two properties indicate that the downstream effects of observation noise are shaped by cooperative dependencies among agents: although disturbances are introduced independently, their impacts become structured through the underlying cooperation structures.

Since cooperation structures influence how information is exchanged and decisions are coupled among agents, a line of research has focused on modeling agent relationships in cooperative MARL.Early MARL approaches mainly relied on the centralized training and decentralized execution (CTDE) paradigm\cite{amato2024introduction}. Although value decomposition like QMIX\citep{rashid2018qmix,son2019qtran}and centralized critic approaches such as MADDPG\cite{lowe2017multi} have achieved remarkable success, the dependencies among agents are usually captured implicitly through learned value functions or centralized representations.

To overcome this limitation, subsequent studies began to explicitly model relationships among agents through learned coordination structures. Role discovery\cite{lhaksmana2018role,wang2020roma} and grouping-based\cite{russell2003q,phan2021vast} approaches identify functional specialization or local cooperation units among agents. More recently, graph-based\citep{malysheva2018deep,jiang2018graph,zang2023automatic} like MAGNet relational models have been widely adopted to capture dynamic interactions among agents. Graph neural networks model pairwise dependencies\cite{wang2021context}, while attention mechanisms enable adaptive interaction weighting. Hypergraph-based\citep{feng2019hypergraph,liu2025hygma} approaches extend these ideas to higher-order interactions among multiple agents. These methods substantially improve the modeling of complex cooperation patterns in MARL. Despite these advances, they mainly focus on learning effective coordination structures and rarely investigate how cooperation structures shape the effects of observation noise.

Other studies focus on investigates robustness learning under noisy observations. Recent studies have explored several strategies to alleviate the adverse effects of observation noise. MADDPG-M\cite{kilinc2018multi} enhances robustness by introducing a communication mechanism that enables agents to access additional information from teammates. However, this approach relies on an additional communication medium, which may not always be available in practical scenarios. \cite{lin2020robustness} investigate the vulnerability of cooperative MARL algorithms under noisy observations, but they do not provide an explicit robustness learning mechanism. More recent approaches improve robustness through single agent perturbations defense \cite{zhang2020robustdeep}, learn from adversarial attack\cite{zhang2021robust}, and observation discretizations\cite{fu2024ldr}. However, they typically model noise as independent uncertainty associated with individual agents and overlook how cooperation relationships among agents influence the propagation and accumulation of noise effects.

To bridge this gap, we propose SIGMA, a hierarchical collaboration framework that explicitly exploits cooperation structures to improve robustness under noisy observations. 
Motivated by the locally correlated nature of structured noise effects, SIGMA first organizes agents into adaptive local structures through density-based grouping, providing meaningful structural units for subsequent aggregation. 
Within each group, high-order interactions among agents are modeled and their representations are aggregated through intra-group consensus, which preserve shared task-relevant information while smoothing agent-specific representation deviations. 
Given the global heterogeneity of structured noise effects, SIGMA further employs inter-group attention to adaptively model dependencies among different groups and integrate group-level information, thereby preserving cross-group coordination.
Through this local-to-global collaboration process, SIGMA learns robust cooperative representations while preserving both local structural information and global task coordination.

The main contributions of this work are summarized as follows:
\begin{itemize}
    \item We characterize \emph{structured noise effects} in cooperative MARL, revealing that independently introduced observation disturbances can induce locally correlated and globally heterogeneous effects on cooperative decision-making through underlying cooperation structures.
    \item We propose SIGMA, a hierarchical collaboration framework that exploits structured noise effects through adaptive grouping and local-to-global representation aggregation. 
SIGMA consolidates cooperative information within adaptive local structures and further models dependencies across groups, enabling robust representation learning while preserving global coordination.
    \item We conduct extensive experiments on noisy-observation tasks in SMAC to empirically validate the proposed structured noise effects and evaluate the robustness of SIGMA. 
The results demonstrate consistent robustness improvements under observation noise while maintaining competitive performance in noise-free environments.
\end{itemize}

\section{Methodology}

\subsection{Problem Formulation}

We consider a cooperative multi-agent reinforcement learning (MARL) problem with noisy observations, which can be formulated as a decentralized partially observable Markov decision process (Dec-POMDP). The environment is defined as:

\begin{equation}
\mathcal{M}=
\left\langle
\mathcal{N},
\mathcal{S},
\mathcal{A},
P,
R,
\Omega,
\gamma
\right\rangle ,
\end{equation}

where $\mathcal{N}=\{1,\ldots,N\}$ denotes the set of agents, 
$\mathcal{S}$ represents the global state space, 
$\mathcal{A}=\{A_{1},\ldots,A_{N}\}$ denotes the joint action space, 
$P$ represents the state transition function, 
$R$ denotes the shared reward function, 
$\Omega=\{\Omega_{1},\ldots,\Omega_{N}\}$ represents the observation spaces, 
and $\gamma$ is the discount factor.

At timestep $t$, each agent $i$ receives a local observation generated from the underlying environment state:

\begin{equation}
o_i^t=O_i(s^t),
\end{equation}

where $s^t\in\mathcal{S}$ denotes the global state. 
However, real-world environments inevitably introduce observation uncertainties due to sensing errors, communication limitations, and environmental disturbances. Therefore, the actual observation received by agent $i$ is formulated as:

\begin{equation}
\widetilde{o}_i^t=o_i^t+\epsilon_i^t,
\end{equation}

where $\epsilon_i^t$ denotes the observation disturbance. 
Following common robust MARL settings, observation noise is assumed to be independently sampled across agents, i.e.,

\begin{equation}
\epsilon_i^t \perp \epsilon_j^t,\quad i\neq j .
\end{equation}

The objective is to learn decentralized policies

\begin{equation}
\pi=\{\pi_1,\ldots,\pi_N\},
\end{equation}

that maximize the expected cumulative team reward:

\begin{equation}
J(\pi)
=
\mathbb{E}_{\pi}
\left[
\sum_{t=0}^{T}
\gamma^t r^t
\right],
\end{equation}

while maintaining robust cooperative behaviors under noisy observations.

Although observation disturbances are independently introduced to individual agents, independence at the observation level does not necessarily imply independent impacts on cooperative decision-making. In fully cooperative MARL, agents share the same team objective, while their individual decisions jointly contribute to the cumulative team reward. Meanwhile, the interaction dependencies underlying such cooperation are generally non-uniform: agents may differ in their task roles, local interactions, and contributions to different components of the shared objective. Consequently, independently introduced observation disturbances may induce cooperation-dependent effects on agent decision-making.

This distinction motivates us to investigate how observation disturbances interact with the latent cooperation structure among agents. In the following, we formalize these cooperation-dependent disturbance impacts as \emph{structured noise effects}, which provide the motivation for the subsequent structure-aware representation learning framework.

\subsection{Structured Noise Effects Analysis}

Although observation disturbances are independently introduced to individual agents, their induced impacts on cooperative decision-making are not necessarily independent. In fully cooperative MARL, agents share a common team objective, while their individual decisions jointly contribute to the team return. However, such global cooperation does not imply uniform interaction dependencies among all agents. Depending on their task roles, local interactions, and behavioral coordination, different agents may exhibit different degrees of dependency during task execution. Consequently, independently introduced observation disturbances can produce structure-dependent impacts through these underlying cooperative interactions.

To characterize such impacts without restricting the analysis to a specific MARL architecture, let $\Phi_i^t$ denote the decision-related output of agent $i$ at timestep $t$. Given the same underlying environment trajectory, the corresponding outputs under clean and noisy observations are defined as

\begin{equation}
\Phi_i^{\mathrm{clean},t}
=
\Phi_i(o_i^{0:t}),
\qquad
\Phi_i^{\mathrm{noisy},t}
=
\Phi_i(\widetilde{o}_i^{0:t}).
\end{equation}

where $\Phi_i^t$ may correspond to local action-value estimates in value-based methods or policy outputs in policy-based methods. The noise-induced decision impact of agent $i$ is then characterized by

\begin{equation}
E_i^t
=
\mathcal{D}
\left(
\Phi_i^{noisy,t},
\Phi_i^{clean,t}
\right),
\end{equation}

where $\mathcal{D}(\cdot,\cdot)$ denotes a general discrepancy measure between the clean and noisy decision outputs. This formulation distinguishes the independently introduced observation disturbance $\epsilon_i^t$ from its downstream impact $E_i^t$ on cooperative decision-making. Therefore, independence among observation disturbances does not necessarily imply independence or uniformity among their induced decision impacts.

Based on this distinction, we characterize \emph{structured noise effects} from two complementary perspectives: \emph{local correlation} and \emph{global heterogeneity}.

\textit{Local correlation.}
Although all agents contribute to the same global objective, their task-related interaction dependencies are generally non-uniform. Some agents may be more strongly coupled through their local interactions and coordinated behaviors, whereas others exhibit relatively weak dependencies. Such differences can further shape how independently introduced observation disturbances affect their cooperative decisions.

Let $\mathcal{P}_{\mathrm{strong}}$ and $\mathcal{P}_{\mathrm{weak}}$ denote agent pairs with relatively strong and weak task-related interaction dependencies, respectively. We hypothesize that noise-induced decision impacts exhibit stronger temporal correlations among strongly interacting agents:

\begin{equation}
\mathbb{E}_{(i,j)\in\mathcal{P}_{\mathrm{strong}}}
\left[\rho(E_i,E_j)\right]
>
\mathbb{E}_{(i,j)\in\mathcal{P}_{\mathrm{weak}}}
\left[\rho(E_i,E_j)\right].
\end{equation}

Importantly, this correlation does not imply that the original observation disturbances become statistically correlated. Instead, it reflects the cooperation-dependent responses of agents to independently introduced disturbances. We refer to this property as the local correlation of structured noise effects.

\textit{Global heterogeneity.}
In addition to local correlation, noise-induced decision impacts may be distributed non-uniformly across the multi-agent system. Different agents can have different task roles, interaction contexts, and contributions to the shared objective, and may therefore exhibit different sensitivities to observation disturbances. We characterize the overall disturbance sensitivity of agent $i$ over a trajectory as

\begin{equation}
H_i
=
\frac{1}{T}
\sum_{t=1}^{T}
E_i^t .
\end{equation}

The resulting sensitivities are not necessarily uniform across agents. Moreover, when agents exhibit different task-related interaction structures, such heterogeneity can also emerge at the level of cooperative groups. Therefore, structured noise effects may simultaneously exhibit locally correlated variations among strongly interacting agents and heterogeneous impact intensities across the broader cooperation structure.

Together, these two properties characterize the structured nature of observation-noise effects in cooperative MARL: noise-induced decision impacts tend to be locally correlated according to task-related interaction dependencies, while remaining globally heterogeneous across agents and cooperation structures. This observation suggests that robust cooperative decision-making should not only address observation uncertainty at the individual-agent level, but also explicitly account for the latent cooperation structures through which disturbance impacts are manifested.

\subsection{Framework Overview}

The central motivation of SIGMA is that independently introduced observation disturbances can induce structured impacts on cooperative decision-making. 
As characterized above, these structured noise effects exhibit two complementary properties: noise-induced decision impacts tend to be locally correlated among agents with stronger task-related dependencies, while their influence remains globally heterogeneous across different agents and local structures. 
Therefore, robust cooperative learning under noisy observations requires not only handling individual observation uncertainty, but also exploiting the underlying local structures through which disturbance effects are organized.

To address this challenge, SIGMA introduces a hierarchical collaboration framework that organizes agents into adaptive local structures and progressively integrates cooperative information from local to global levels. 
Rather than directly aggregating information across the entire agent population, SIGMA first identifies local structural units according to the similarity of agent observations. 
It then performs consensus aggregation within each group to consolidate locally shared information and alleviate agent-specific disturbances. 
Finally, interactions among different groups are modeled to recover global cooperative dependencies while accounting for their heterogeneous roles. 
In this way, SIGMA combines local structural aggregation with global coordination to improve cooperative representation learning under noisy observations.

As illustrated in Fig.~\ref{fig:framework}, SIGMA consists of three main components:
\begin{figure*}[t]
    \centering
    \includegraphics[width=0.82\textwidth]{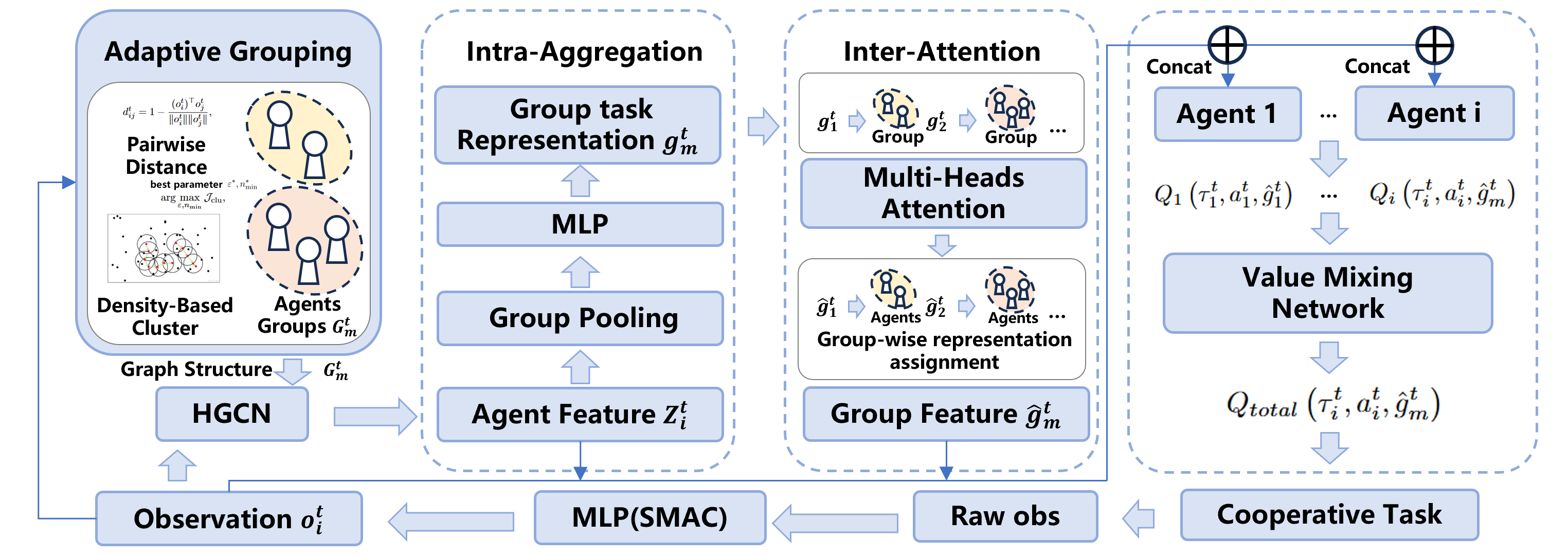}
    \caption{Overall framework of SIGMA for robust cooperative representation learning under noisy observations.}
    \label{fig:framework}
\end{figure*}

\textbf{Adaptive Grouping.}
Motivated by the locally correlated nature of structured noise effects, SIGMA first identifies local structural units before hierarchical aggregation. 
Since task-related dependencies are latent and may evolve with task states, the module dynamically groups agents according to the similarity of their local observations. 
The resulting adaptive groups define the structural boundaries for subsequent intra-group consensus aggregation.

\textbf{Intra-group Consensus Aggregation.}
Within each adaptive group, SIGMA models high-order interactions among agents and performs consensus aggregation over their interaction-enhanced representations. 
This process consolidates information shared within the local structure while smoothing agent-specific representation disturbances. 
By restricting consensus aggregation to adaptive groups, SIGMA avoids indiscriminately mixing information across structurally different agents and constructs robust group-level cooperative representations.

\textbf{Inter-group Attention.}
The locally aggregated groups still need to exchange information to support global cooperation, while the global heterogeneity of structured noise effects indicates that different groups should not necessarily contribute equally. 
SIGMA therefore employs inter-group attention to adaptively model dependencies among group-level representations. 
This mechanism restores cross-group information exchange while preserving heterogeneous group-level contributions, producing a global task representation for decentralized decision-making.

Overall, SIGMA follows a hierarchical collaboration paradigm: agents are first organized into adaptive local structures, cooperative information is then consolidated within each group, and the resulting group-level representations are finally integrated across groups to recover global coordination. 
Through this local-to-global architecture, SIGMA exploits the structural characteristics of noise-induced decision impacts while preserving task-relevant cooperative information, thereby supporting robust decentralized decision-making under noisy observations.

\subsection{Adaptive Grouping}

The local correlation of structured noise effects suggests that noise-induced impacts are associated with local task-related dependencies rather than being uniformly distributed across all agents. 
This motivates SIGMA to identify local structural units before performing hierarchical aggregation. 
However, such task-related dependencies are latent and may dynamically evolve with task states. 
Therefore, instead of relying on predefined team assignments, SIGMA constructs adaptive groups according to the similarity of agents' local observations.

At timestep $t$, agents operating under related local task contexts tend to exhibit similar observation patterns. 
SIGMA therefore uses observation similarity as a practical criterion for identifying local structural relationships. 
Specifically, the pairwise distance between agents $i$ and $j$ is defined as

\begin{equation}
d_{ij}^t=
1-
\frac{
(o_i^t)^{\top}o_j^t
}{
\|o_i^t\|\|o_j^t\|
},
\end{equation}

where a smaller $d_{ij}^t$ indicates greater similarity between the local task contexts perceived by the two agents. Based on the resulting pairwise distance matrix, SIGMA employs a dynamically parameterized density-based clustering procedure to construct cooperative groups.

A key challenge of density-based clustering is that its grouping results depend on the neighborhood radius $\varepsilon$ and the minimum number of neighboring samples $n_{\min}$. Fixed clustering parameters may be unsuitable for MARL because the distribution of agent observations continuously changes during task execution. To accommodate such variations, SIGMA dynamically searches for appropriate clustering parameters according to the current observation distribution.

Specifically, SIGMA first generates several candidate neighborhood radii from the current distribution of pairwise observation distances. 
Instead of using fixed radius values, different quantiles of the distance distribution are selected so that the candidate radii can adapt to the changing observation patterns of agents:

\begin{equation}
\mathcal{E}^{t}
=
\left\{
Q_{q}\left(D^{t}\right)
\mid
q\in\mathcal{Q}
\right\},
\end{equation}

where $D^{t}$ contains all pairwise observation distances at timestep $t$, 
$Q_q(\cdot)$ denotes the $q$-th quantile of these distances, and 
$\mathcal{Q}$ specifies the quantiles used to generate candidate radii. 
Each candidate radius $\varepsilon\in\mathcal{E}^{t}$ is then combined with candidate values of the minimum neighborhood size $n_{\min}$, and DBSCAN systematically evaluates all candidate combinations of $\varepsilon$ and $n_{\min}$ , thereby obtaining a set of candidate grouping results. 

To select an appropriate grouping from the candidate clustering results, 
SIGMA evaluates each candidate from three complementary aspects: 
\emph{structural quality}, \emph{agent coverage}, and \emph{grouping granularity}. 
A desirable grouping should form compact and well-separated groups, retain as many agents as possible in meaningful cooperative groups, and avoid both excessive merging and fragmentation. 
Accordingly, the clustering score is defined as

\begin{equation} 
\mathcal{J}_{\mathrm{clu}} 
= 
S_{\mathrm{sil}} 
- 
\lambda_{n}R_{\mathrm{noise}} 
- 
\lambda_{g}R_{\mathrm{group}}, 
\end{equation} 

where $S_{\mathrm{sil}}$ denotes the silhouette score and favors groupings with high intra-group similarity and clear inter-group separation. 
However, relying solely on the silhouette score may favor overly restrictive partitions that leave many agents unassigned. 
Therefore, $R_{\mathrm{noise}}$ penalizes candidate groupings containing excessive noise points, encouraging sufficient agent coverage. 
Meanwhile, $R_{\mathrm{group}}$ regulates the grouping granularity by penalizing undesirable deviations in the number of discovered groups. 
This prevents the grouping structure from collapsing most agents into a few overly broad groups or fragmenting them into excessively small groups. 
The coefficients $\lambda_n$ and $\lambda_g$ control the strengths of the two penalties.

By jointly considering these three criteria, the evaluation favors groupings that are structurally distinguishable, sufficiently inclusive, and appropriately partitioned for subsequent intra-group aggregation.

The clustering configuration with the highest score is selected:

\begin{equation}
(\varepsilon^{*},n_{\min}^{*})
=
\arg\max_{\varepsilon,n_{\min}}
\mathcal{J}_{\mathrm{clu}},
\end{equation}

and the corresponding grouping result is denoted as

\begin{equation}
\mathcal{G}^{t}
=
\{
G_1^t,G_2^t,\ldots,G_M^t
\}.
\end{equation}

Since DBSCAN may identify isolated agents as noise points, SIGMA assigns each such agent to an individual singleton group rather than discarding it from subsequent cooperative representation learning. Therefore, every agent remains explicitly represented in the hierarchical aggregation process.

To avoid excessive fluctuations in the grouping structure, group assignments are updated periodically rather than at every timestep. A newly obtained grouping is adopted only when it satisfies the predefined stability criterion; otherwise, the previous grouping is retained. This temporal stabilization prevents transient observation variations from causing frequent changes in cooperative group assignments.

Overall, the adaptive grouping procedure determines both the grouping structure and its density parameters according to the evolving distribution of agent observations. This enables SIGMA to construct flexible local structural units without requiring a predefined number of cooperative groups, providing the basis for subsequent intra-group consensus aggregation.

\subsection{Intra-group Consensus Aggregation}

After identifying adaptive groups, SIGMA further performs intra-group consensus aggregation to construct robust group-level cooperative representations. 
The adaptive grouping provides local structural units composed of agents with related task contexts, within which agents are more likely to share task-relevant information while exhibiting locally associated responses to observation disturbances. 
However, individual agent representations may still contain deviations arising from agent-specific information, partial observability, and observation uncertainties. 
Therefore, rather than treating agents independently or directly mixing information across the entire agent population, SIGMA consolidates representations within each adaptive group to extract locally shared cooperative information while reducing the influence of individual representation deviations.

For each adaptive group $G_m^t$, SIGMA first employs a hypergraph neural network (HGCN) to encode high-order interactions among agents within the group:

\begin{equation}
z_i^t=
\mathrm{HGCN}(o_i^t,G_m^t),
\quad i\in G_m^t ,
\end{equation}

where $z_i^t$ denotes the interaction-enhanced representation of agent $i$. 
The hypergraph formulation captures collective dependencies among multiple agents within the same local structure, providing interaction-aware representations for subsequent consensus aggregation.

To illustrate the information shared within an adaptive group, the representation of each agent can be conceptually decomposed into a group-shared component and an individual deviation:

\begin{equation}
z_i^t=s_m^t+\delta_i^t,
\quad i\in G_m^t ,
\end{equation}

where $s_m^t$ represents the semantic information shared among agents in group $G_m^t$, and $\delta_i^t$ denotes the agent-specific deviation from the shared component. 
Such deviations may arise from differences in local information, partial observability, and observation disturbances.

Based on these interaction-aware representations, SIGMA performs consensus aggregation within each adaptive group:

\begin{equation}
g_m^t=
\frac{1}{|G_m^t|}
\sum_{i\in G_m^t}z_i^t
=
s_m^t+
\frac{1}{|G_m^t|}
\sum_{i\in G_m^t}\delta_i^t ,
\end{equation}

where $g_m^t$ denotes the resulting group-level cooperative representation. 
As shown in the above formulation, intra-group aggregation preserves the shared component $s_m^t$ while averaging agent-specific deviations within the local structure. 
This consensus process smooths individual representation fluctuations and reduces the influence of any single agent on the group-level representation.

Under structured noise effects, however, agent deviations may contain both agent-specific and locally correlated components. 
While agent-specific disturbances can be attenuated through aggregation, locally correlated disturbances may persist within the group, limiting the noise-reduction capability of simple averaging. 
Therefore, intra-group consensus provides a structured way to suppress individual disturbances while retaining information shared within each local cooperative structure.

Consequently, SIGMA obtains a set of group-level representations
$\{g_1^t,\ldots,g_M^t\}$, each characterizing a distinct local cooperative structure. 
Rather than prematurely merging these representations, SIGMA retains their structural differences for subsequent inter-group coordination.

\subsection{Inter-group Attention}

After intra-group consensus aggregation, SIGMA obtains multiple group-level representations that characterize different local cooperative structures. 
While such local aggregation facilitates robust representation learning within each group, it may also limit information exchange across different groups. 
Since cooperative tasks generally require coordination beyond individual local structures, SIGMA further models inter-group dependencies to restore global cooperative interactions.

Meanwhile, the global heterogeneity of structured noise effects suggests that different groups may exhibit distinct task characteristics and disturbance sensitivities. 
Therefore, simply treating all groups equally may overlook their heterogeneous roles in global cooperation. 
To address both aspects, SIGMA employs an inter-group attention mechanism to adaptively model dependencies among different groups and integrate their information into a global task representation.

Given the group-level representations $\{g_m^t\}_{m=1}^{M}$, SIGMA employs an inter-group attention mechanism to adaptively model dependencies among different local cooperative structures.

Specifically, the query, key, and value representations of each group are computed as

\begin{equation}
q_m^t=W_q g_m^t,\qquad
k_m^t=W_k g_m^t,\qquad
v_m^t=W_v g_m^t,
\end{equation}

where $W_q$, $W_k$, and $W_v$ are learnable projection matrices.

The attention coefficient between groups $m$ and $n$ is calculated as

\begin{equation}
\alpha_{mn}^t
=
\frac{
\exp((q_m^t)^\top k_n^t/\sqrt{d})
}
{
\sum_{j=1}^{M}
\exp((q_m^t)^\top k_j^t/\sqrt{d})
},
\end{equation}

where $d$ denotes the dimension of the key representation. 
The attention coefficient $\alpha_{mn}^t$ characterizes the context-dependent relevance of group $n$ to group $m$, allowing SIGMA to distinguish heterogeneous inter-group dependencies rather than treating all groups equally.

Based on these attention weights, each group representation is refined by integrating information from other cooperative groups:

\begin{equation}
\hat{g}_m^t
=
\sum_{n=1}^{M}
\alpha_{mn}^t v_n^t .
\end{equation}

Through this interaction, locally aggregated group representations can exchange information across structural boundaries, recovering cross-group cooperative dependencies that may be weakened by intra-group aggregation. Meanwhile, the adaptive attention weights allow each group to selectively incorporate information from other groups according to the current task context, while preserving the distinctions among different cooperative structures.

As a result, each group obtains a refined representation $\hat{g}_m^t$ that retains its group-specific cooperative information while incorporating relevant context from other groups. Rather than collapsing all groups into a single global representation, SIGMA maintains these refined representations separately and assigns $\hat{g}_m^t$ to the agents belonging to group $G_m^t$ for subsequent decentralized decision-making.

Together with intra-group consensus aggregation, inter-group attention therefore enables SIGMA to preserve locally shared cooperative information while restoring coordination across different cooperative structures. This forms a hierarchical representation process from individual agents to local cooperative groups and finally to group-specific representations enriched with global cooperative
context.

\subsection{Integration with Cooperative MARL Frameworks}

SIGMA is designed as a representation learning module that can be integrated into existing cooperative MARL frameworks. Instead of modifying the policy optimization procedure, SIGMA focuses on learning robust cooperative representations from noisy observations and provides additional cooperative information for decentralized decision-making.

Specifically, after inter-group attention, each cooperative group obtains a refined representation $\hat{g}_m^t$ that incorporates cross-group contextual information while retaining its group-specific cooperative information.
For each agent $i\in G_m^t$, the corresponding group representation $\hat{g}_m^t$ is assigned to the agent and incorporated into its decision function together with its local information:

\begin{equation}
\pi_i
\left(
a_i^t
\mid
\tau_i^t,\hat{g}_m^t
\right),
\quad i\in G_m^t ,
\end{equation}

where $\tau_i^t$ denotes the local action-observation history of agent $i$, and $\hat{g}_m^t$ denotes the refined cooperative representation of the adaptive group to which agent $i$ belongs.

For value-based MARL algorithms, the corresponding group representation can similarly be incorporated into the individual value estimation:

\begin{equation}
Q_i
\left(
\tau_i^t,a_i^t,\hat{g}_m^t
\right),
\quad i\in G_m^t .
\end{equation}

In this way, agents within the same adaptive group share the corresponding refined cooperative representation, while agents in different groups receive group-specific representations enriched with cross-group cooperative context. Therefore, SIGMA can be integrated into existing cooperative MARL algorithms
as an additional representation learning module without modifying their underlying policy optimization procedures.

\section{Experiments}

\subsection{Experimental Setup}

\textbf{Environment.}
We evaluate SIGMA on the StarCraft II Multi-Agent Challenge (SMAC), a widely used benchmark for cooperative multi-agent reinforcement learning.
Experiments are conducted on the \texttt{5m\_vs\_6m} and \texttt{8m\_vs\_9m} scenarios, where multiple allied agents must coordinate their movements, target selection, and combat behaviors against an opposing team.
These scenarios provide representative cooperative settings for evaluating both policy performance and robustness under imperfect observations.

\textbf{Noise Settings.}
To evaluate robustness against observation uncertainty, Gaussian noise is independently introduced into the local observation of each agent:
\begin{equation}
\widetilde{o}_i^t
=
o_i^t+\epsilon_i^t,
\qquad
\epsilon_i^t\sim\mathcal{N}(0,\sigma^2),
\end{equation}
where $\sigma$ controls the observation-noise intensity.
We consider both the noise-free setting and multiple noisy settings with
$\sigma\in\{0,1.5,2.5,5.0\}$.
Importantly, observation disturbances are independently sampled across agents, allowing us to investigate whether independent input disturbances can induce structured downstream effects through cooperative interactions.

\textbf{Baselines.}
We compare SIGMA with representative cooperative MARL methods, including QMIX and HYGMA.
QMIX serves as a widely adopted value-decomposition baseline, while HYGMA models structured interactions among multiple agents through dynamic grouping and hypergraph-based coordination.
All methods are evaluated under the same environment and observation-noise settings to ensure a consistent comparison.

\textbf{Implementation Details.}
For each scenario, all methods are trained using the same environment configurations and evaluation protocol.
Unless otherwise specified, the converged checkpoint obtained at the end of training is used for robustness evaluation and structured noise analysis, without further policy updates during evaluation.
Other training hyperparameters and implementation details are kept consistent across compared settings whenever applicable.

\textbf{Evaluation Metrics.}
We primarily use episode win rate to evaluate cooperative policy performance under different observation-noise levels, while training curves are used to examine the learning dynamics of different methods.
For the analysis of structured noise effects, we further evaluate two complementary properties at the Q-value level.
Local correlation is characterized by the temporal correlation of relative noise-induced Q-effects among different agent pairs, whereas global heterogeneity is quantified using the normalized disparity of centered Q-effects across adaptive cooperative groups.
The detailed construction of these analysis metrics and their corresponding structural controls is introduced in the structured noise analysis below.

\subsection{Performance Comparison under Noisy Observations}

We first evaluate the overall robustness of SIGMA under different levels of observation noise.
Table~\ref{tab:main_results} reports the final win rates of SIGMA and the compared methods under both noise-free and noisy conditions.

\begin{table}[t]
\centering
\caption{Performance comparison under different observation-noise levels on SMAC.}
\label{tab:main_results}

\resizebox{\linewidth}{!}{
\begin{tabular}{llcccc}
\toprule
Scenario & Method & $\sigma=0$ & $\sigma=1.5$ & $\sigma=2.5$ & $\sigma=5.0$ \\
\midrule

\multirow{3}{*}{\texttt{5m\_vs\_6m}}
& QMIX & 79.87\%& 35.64\%& 40.00\%& 11.87\%\\
& HYGMA & \textbf{96.88\%}& 88.75\%& 83.13\%& 0.25\%\\
& SIGMA (Ours) & 95.63\%& \textbf{94.37\%}& \textbf{90.00\%}& \textbf{66.25\%}\\

\midrule

\multirow{3}{*}{\texttt{8m\_vs\_9m}}
& QMIX & 83.13\%& 82.26\%& 65.89\%& 16.25\%\\
& HYGMA & \textbf{98.45\%}& 88.12\%& 77.50\%& 23.75\%\\
& SIGMA (Ours) & 97.24\%& \textbf{95.00\%}& \textbf{91.87\%}& \textbf{94.37\%}\\

\bottomrule
\end{tabular}
}

\vspace{1mm}
\footnotesize
$\sigma$ denotes the standard deviation of observation noise.
\end{table}

As shown in Table~\ref{tab:main_results}, the performance differences among the compared methods become increasingly pronounced as the observation-noise intensity increases.
While all methods achieve competitive performance under noise-free or mild-noise conditions, QMIX suffers substantial degradation under stronger disturbances.
The structure-aware baseline HYGMA generally provides improved robustness by explicitly modeling multi-agent interactions, but its performance also decreases as observation noise becomes stronger.
In contrast, SIGMA maintains consistently high win rates across the evaluated noise levels and scenarios.
These results suggest that explicitly modeling cooperation structures alone does not necessarily provide sufficient robustness against noisy observations.
The advantage of SIGMA lies in further exploiting these structures according to the characteristics of structured noise effects.
Specifically, intra-group consensus integrates information among structurally related agents to reduce the influence of agent-specific deviations, while inter-group attention preserves and adaptively coordinates information across groups with different disturbance responses.
Consequently, the performance advantage of SIGMA becomes more evident as local observations become increasingly unreliable.

\begin{figure}[t]
    \centering
    \begin{subfigure}[t]{0.49\linewidth}
        \centering
        \includegraphics[width=\linewidth]{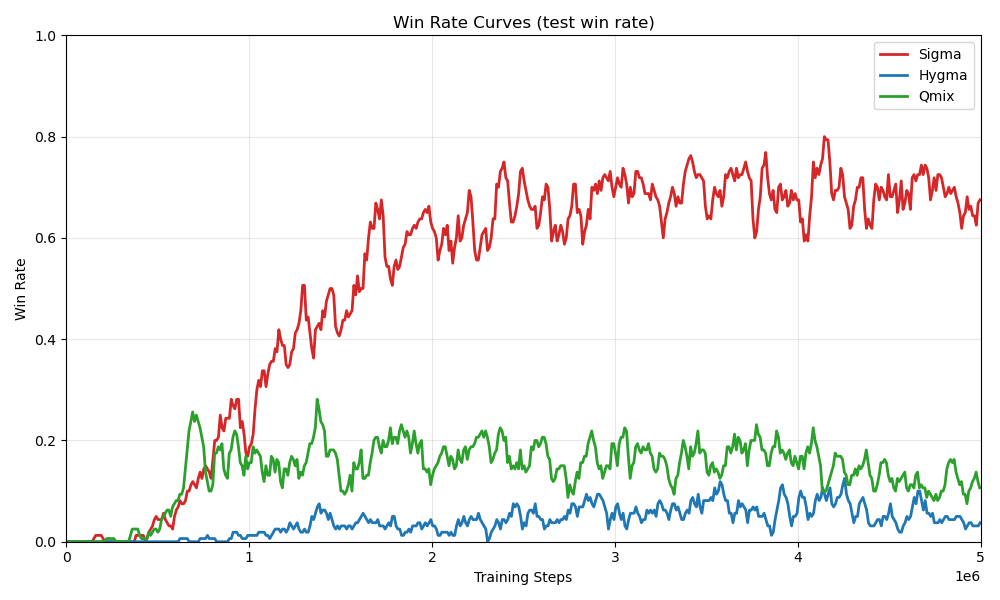}
        \caption{\texttt{5m\_vs\_6m}}
        \label{fig:win_rate_5m6m}
    \end{subfigure}
    \hfill
    \begin{subfigure}[t]{0.49\linewidth}
        \centering
        \includegraphics[width=\linewidth]{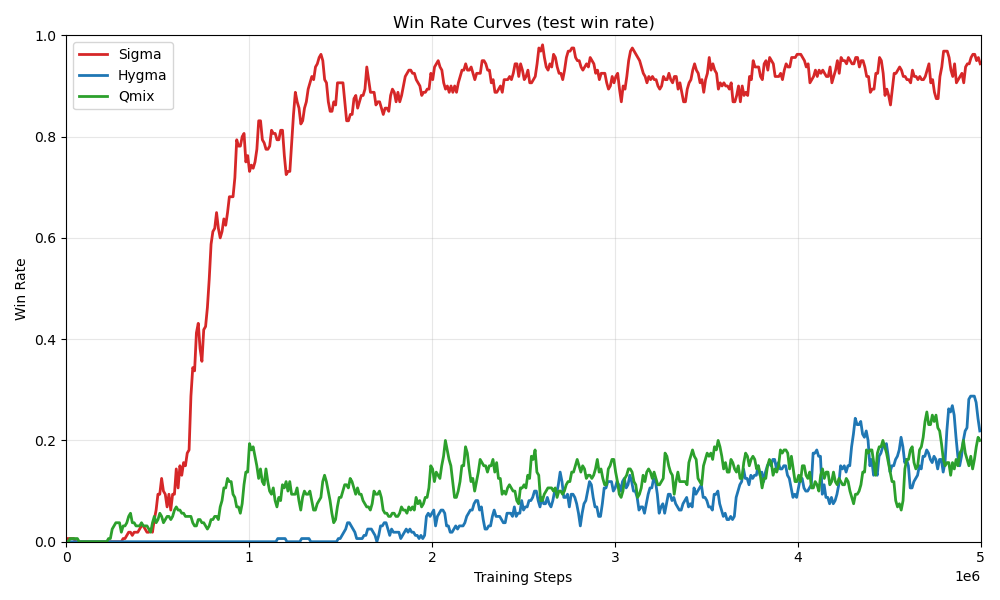}
        \caption{\texttt{8m\_vs\_9m}}
        \label{fig:win_rate_8m9m}
    \end{subfigure}
    \caption{Training performance of different methods under strong observation noise ($\sigma=5.0$) on two SMAC scenarios.}
    \label{fig:win_rate}
\end{figure}

Figure~\ref{fig:win_rate} further compares the learning dynamics of different methods under strong observation noise.
Rather than considering only the final policy performance, the training curves illustrate how effectively each method learns and maintains cooperative behaviors throughout training.

As shown in Fig.~\ref{fig:win_rate}, SIGMA achieves a faster improvement in win rate and maintains more stable performance during training compared with the baseline methods.
The advantage becomes increasingly evident as training progresses, suggesting that exploiting adaptive cooperation structures facilitates more effective policy learning when local observations are disturbed.
Together with the final performance results in Table~\ref{tab:main_results}, these results demonstrate that SIGMA improves both the robustness and learning efficiency of cooperative policies under noisy observations.

\subsection{Analysis of Structured Noise Effects}

The design of SIGMA is motivated by the \emph{structured noise effects}, where independently introduced observation disturbances may induce locally correlated yet globally heterogeneous effects on cooperative decision-making.
To empirically examine this phenomenon, we analyze noise-induced changes in agent Q-values from two complementary perspectives: local correlation among agents and global heterogeneity across cooperative groups.

To isolate the effects induced by observation disturbances, we perform paired clean/noisy forward passes from the same decision state.
At each timestep, the clean and noisy branches share the same environment state, observation history, hidden state, and available-action set, while observation noise is introduced only into the noisy branch.
The resulting Q-value differences therefore characterize the downstream decision effects associated with observation disturbances.
For structural comparison, we further construct random groupings by randomly reassigning agents while preserving the number and sizes of the adaptive groups.
This provides a control for determining whether the observed noise effects are specifically associated with the discovered cooperation structures rather than arbitrary partitions.

\textbf{Local Correlation.}
We first examine whether agents within the same adaptive group exhibit more correlated noise-induced decision effects than agents belonging to different groups.
For agent $i$, we characterize the relative Q-effect at timestep $t$ as the normalized change between its clean and noisy Q-values:

\begin{equation}
E_{i,\mathrm{rel}}^t
=
\frac{
\left\|
Q_{i,\mathrm{noisy}}^t
-
Q_{i,\mathrm{clean}}^t
\right\|_2
}{
\left\|
Q_{i,\mathrm{clean}}^t
\right\|_2+\epsilon
},
\end{equation}

where $\epsilon$ is a small constant for numerical stability.
This metric measures the magnitude of the noise-induced Q-value change relative to the agent's original Q-value scale.

We then compute the temporal correlation of relative Q-effects for different agent pairs.
Let $\rho_{\mathrm{within}}$ and $\rho_{\mathrm{between}}$ denote the average correlations for agent pairs within the same adaptive group and across different groups, respectively.
Their difference is defined as

\begin{equation}
\Delta\rho
=
\rho_{\mathrm{within}}
-
\rho_{\mathrm{between}}.
\end{equation}

A positive $\Delta\rho$ indicates that agents within the same local cooperation structure exhibit more strongly correlated noise effects.

Table~\ref{tab:local_corr} reports the local-correlation results over five repeated runs on \texttt{5m\_vs\_6m}.
For comparison, we also report the correlation gap obtained from the matched random groupings.

\begin{table}[t]
\centering
\caption{Local correlation of relative Q-effects across SMAC scenarios.}
\label{tab:local_corr}

\resizebox{\linewidth}{!}{
\begin{tabular}{lcccc}
\toprule
Scenario & Within & Between & $\Delta\rho$ & Random $\Delta\rho$ \\
\midrule
\texttt{5m\_vs\_6m}
& $0.308\pm0.012$& $0.076\pm0.004$& $\mathbf{0.232\pm0.009}$& $-0.006\pm0.001$\\
\texttt{8m\_vs\_9m}
& $0.346\pm0.001$& $0.265\pm0.003$& $\mathbf{0.082\pm0.001}$& $0.002\pm0.000$\\
\bottomrule
\end{tabular}
}

\vspace{1mm}
\parbox{\linewidth}{\footnotesize
\textit{Note:} Results are reported as mean $\pm$ standard deviation over five repeated analyses.
}
\end{table}

As shown in Table~\ref{tab:local_corr}, within-group agent pairs exhibit stronger correlations in their relative Q-effects than between-group pairs in both senarios. On \texttt{5m\_vs\_6m}, the average correlation gap $\Delta\rho$ reaches $0.232$, while the corresponding gap under random grouping remains close to zero ($-0.006$). A consistent pattern is observed on \texttt{8m\_vs\_9m}, with an average $\Delta\rho$ of $0.082$, compared with only $0.002$ under random grouping. Moreover, the within-group correlation remains higher than the between-group correlation across all five repeated analyses in both scenarios. These results indicate that the local correlation of noise-induced decision effects is consistently associated with the discovered cooperation structures rather than arbitrary agent partitions, providing empirical evidence for the local-correlation property of structured noise effects.

\textbf{Global Heterogeneity.}
Local correlation characterizes whether noise effects vary coherently among structurally related agents, but does not indicate whether different cooperation structures are affected to the same extent.
We therefore further examine the global heterogeneity of noise effects across adaptive groups.

To focus on differential disturbance responses across agents, we remove the common Q-effect component and obtain the centered Q-effect $E_{i,\mathrm{cen}}^t$.
For each adaptive group $G_m^t$, the corresponding group-level effect is computed as

\begin{equation}
E_m^t
=
\frac{1}{|G_m^t|}
\sum_{i\in G_m^t}
E_{i,\mathrm{cen}}^t .
\end{equation}
We quantify global heterogeneity using the normalized average pairwise difference between group-level effects:

\begin{equation}
H_{\mathrm{group}}^t
=
\frac{
\displaystyle
\frac{2}{M(M-1)}
\sum_{m<n}
\left|
E_m^t-E_n^t
\right|
}{
\displaystyle
\frac{1}{M}
\sum_{m=1}^{M}
E_m^t+\epsilon
},
\end{equation}

where $M$ denotes the number of adaptive groups.
A larger $H_{\mathrm{group}}$ indicates stronger differences in noise effects across local cooperation structures.

Table~\ref{tab:global_heterogeneity} compares the heterogeneity measured under adaptive and matched random groupings.
\begin{table}[t]
\centering
\caption{Global heterogeneity of centered Q-effects across SMAC scenarios.}
\label{tab:global_heterogeneity}

\resizebox{\linewidth}{!}{
\begin{tabular}{lcccc}
\toprule
Scenario & Adaptive $H$ & Random $H$ & $\Delta H$ & Positive $\Delta H$ (\%) \\
\midrule
\texttt{5m\_vs\_6m}
& $1.023 \pm 0.001$& $0.771 \pm0.000$& $\mathbf{0.251 \pm0.001}$& $92\%\pm0.003$\\
\texttt{8m\_vs\_9m}
& $1.094 \pm0.002$& $0.804 \pm 0.000$& $\mathbf{0.290 \pm 0.002}$& $93\%\pm0.002$\\
\bottomrule
\end{tabular}
}

\vspace{1mm}
\parbox{\linewidth}{\footnotesize
\textit{Note:} $H$ denotes the normalized group-level heterogeneity of
centered Q-effects. Results are reported as mean $\pm$ standard deviation
over five repeated analyses. Positive $\Delta H$ denotes the percentage of
analyzed episodes satisfying
$H_{\mathrm{adaptive}}>H_{\mathrm{random}}$.
}
\end{table}

As shown in Table~\ref{tab:global_heterogeneity}, adaptive groups exhibit consistently greater heterogeneity in noise-induced Q-effects than matched random groups across both scenarios. On \texttt{5m\_vs\_6m}, the average heterogeneity increases from $0.771$ under random grouping to $1.023$ under adaptive grouping, yielding an average gap of $\Delta H=0.251$. A similar pattern is observed on \texttt{8m\_vs\_9m}, where the average heterogeneity gap reaches $0.290$.

Moreover, positive heterogeneity gaps are observed in approximately $92\%$ and $93\%$ of the analyzed episodes on \texttt{5m\_vs\_6m} and \texttt{8m\_vs\_9m}, respectively. The consistency across scenarios and repeated analyses indicates that
noise-induced decision effects are not uniformly distributed across agent groups, but exhibit systematic differences associated with the discovered cooperation structures.

\subsection{Ablation Study}

To investigate the contribution of each component in SIGMA, we conduct ablation studies by removing or replacing individual modules.

We consider the following variants:

\begin{itemize}
    \item \textbf{SIGMA w/o Grouping}: removes adaptive grouping and directly aggregates information among all agents.
    \item \textbf{SIGMA-Random Group}: replaces adaptive grouping with randomly generated groups while maintaining the same group sizes.
    \item \textbf{SIGMA w/o Intra}: removes intra-group consensus aggregation and uses individual agent representations directly.
    \item \textbf{SIGMA w/o Inter}: removes inter-group attention and directly combines group-level representations.

\end{itemize}
\begin{figure}[t]
    \centering
    \includegraphics[width=\linewidth]{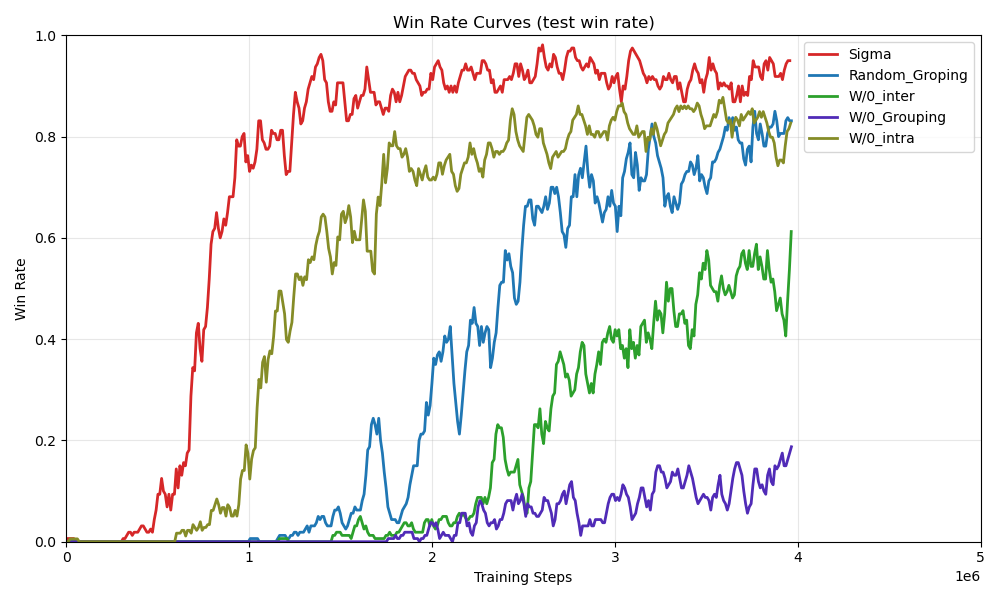}
    \caption{Performance comparison of SIGMA and its ablated variants under noisy observations.}
    \label{fig:ablation}
\end{figure}
Removing adaptive grouping and assigning all agents into a single group results in the largest performance degradation, demonstrating that preserving meaningful cooperation structures is critical for robust representation learning. The random grouping variant achieves better performance than the single-group setting, indicating that group-wise aggregation itself is beneficial; however, its inferior performance compared with SIGMA suggests that adaptively discovering cooperation structures is important for obtaining more effective representations. Removing inter-group attention causes a substantial performance drop, which highlights the importance of modeling dependencies across different cooperative groups for recovering global coordination. In contrast, removing intra-group aggregation leads to a relatively smaller degradation, indicating that intra-group consensus further improves representation robustness by integrating information among related agents.

\subsection{Limitations}

The current experimental evaluation is subject to several limitations.
In particular, the main performance results are currently reported based on individual training runs for each setting. Since MARL training may exhibit variability across random initialization, environment stochasticity, and exploration trajectories, these results should be interpreted as preliminary evidence of the robustness advantage of SIGMA rather than a complete statistical comparison. A more comprehensive evaluation with multiple random seeds, reporting mean performance and standard deviation, will be included in future experiments. In addition, the current evaluation is conducted on a limited set of SMAC scenarios. Future work will extend the evaluation to more diverse cooperative scenarios with different team sizes and coordination complexities to further examine the generalizability of SIGMA.

\section{Conclusion}

In this paper, we investigated the robustness of cooperative MARL under noisy observations from the perspective of cooperation structures. 
We characterized \emph{structured noise effects}, showing that independently introduced observation disturbances can induce decision effects that are locally correlated among structurally related agents while remaining globally heterogeneous across different cooperation structures. 
Our empirical analysis further supports these two properties across different cooperative scenarios, highlighting the role of underlying cooperation structures in shaping the downstream effects of observation noise.

Motivated by this observation, we proposed SIGMA, a hierarchical collaboration framework for robust cooperative representation learning. 
SIGMA adaptively identifies latent cooperation structures, performs intra-group consensus aggregation to integrate information among structurally related agents, and employs inter-group attention to coordinate information across different groups. 
This hierarchical design enables SIGMA to exploit local structural information while preserving dependencies and differences across cooperation structures.

Experiments on the StarCraft II Multi-Agent Challenge show that SIGMA maintains strong cooperative performance under increasing levels of observation noise and exhibits smaller performance degradation than the evaluated baselines. 
Together with the structured noise analysis, these results suggest that explicitly exploiting cooperation structures is a promising approach to improving the robustness of cooperative MARL under noisy observations.

Future work will evaluate SIGMA across a broader range of cooperative tasks and disturbance settings, conduct more comprehensive multi-seed comparisons, and investigate more general mechanisms for modeling and exploiting structured noise effects in large-scale multi-agent systems.
\bibliography{aaai2027}
\end{document}